\documentclass[runningheads]{llncs}
\usepackage{graphicx}
\usepackage{amsmath}
\usepackage{amsfonts}
\usepackage{amstext}
\usepackage{booktabs}
\usepackage{algorithm}
\usepackage{multirow}
\usepackage{epstopdf}
\usepackage{color}
\usepackage{framed}
\usepackage{array}
\usepackage[table,xcdraw]{xcolor}
\usepackage{booktabs}
\usepackage[noend]{algpseudocode}
\usepackage{algorithmicx}
\usepackage{amssymb}
\newcommand{\tabincell}[2]{\begin{tabular}{@{}#1@{}}#2\end{tabular}}

\begin{document}
\title{HIN: Hierarchical Inference Network for Document-Level Relation Extraction}
\titlerunning{HIN}
%
\author{Hengzhu Tang\inst{1,2} \and
Yanan Cao\inst{1} \and
Zhenyu Zhang\inst{1,2} \and
Jiangxia Cao\inst{1,2} \and
Fang Fang\inst{1}\thanks{Corresponding Author}\and
Shi Wang\inst{3} \and
Pengfei Yin\inst{1}
}
%
\authorrunning{H. Tang et al.}
%
\institute{Institute of Information Engineering, Chinese Academy of Sciences,  China \and
School of Cyber Security, University of Chinese Academy of Sciences, China \and
Institute of Computing Technology, Chinese Academy of Sciences, China \\
\email{\{tanghengzhu,caoyanan,zhangzhenyu1996,caojiangxia,fangfang0703,yinpengfei\}@iie.ac.cn} \\
\email{wangshi@ict.ac.cn}
}
\maketitle              
\begin{abstract}
Document-level RE requires reading, inferring and aggregating over multiple sentences.
From our point of view, it is necessary for document-level RE to take advantage of multi-granularity inference information: entity level, sentence level and document level. 
Thus, how to obtain and aggregate the inference information with different granularity is challenging for document-level RE, 
which has not been considered by previous work.
In this paper, we propose a Hierarchical Inference Network (HIN) to make full use of the abundant information from entity level, sentence level and document level.
Translation constraint and bilinear transformation are applied to target entity pair in multiple subspaces to get entity-level inference information.
Next, we model the inference between entity-level information and sentence representation to achieve sentence-level inference information.
Finally, a hierarchical aggregation approach is adopted to obtain the document-level inference information.
In this way, our model can effectively aggregate inference information from these three different granularities.
Experimental results show that our method achieves state-of-the-art performance on the large-scale DocRED dataset.
We also demonstrate that using BERT representations can further substantially boost the performance.

\end{abstract}
\section{Introduction}

\begin{figure}[ht]
	\centering
 	\includegraphics[scale=0.48]{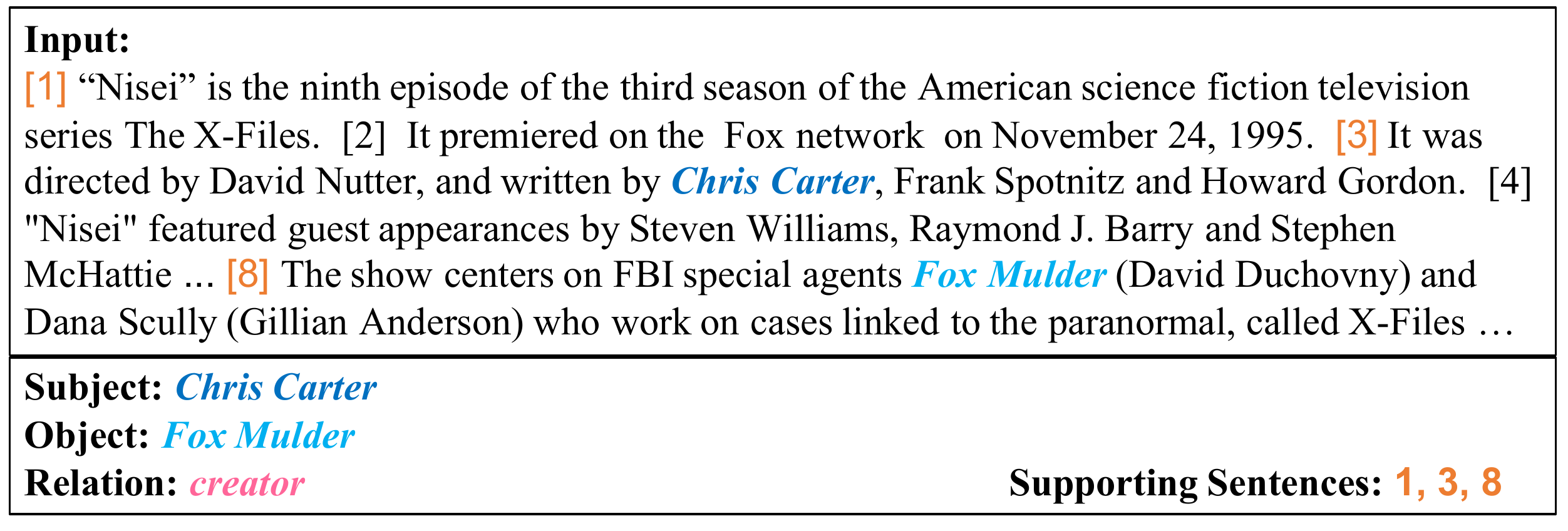}
	\protect\caption{An example from DocRED. Each document in DocRED is annotated with named entity mentions, coreference information, relations, and supporting sentences.
	}
	\label{Fig. 1}
\end{figure}
Relation extraction (RE) aims to detect the semantic relation between entities in plain text, which plays an important role in knowledge base population and natural language understanding.
Most previous work focuses on sentence-level RE, i.e., extracting relational facts from a single sentence.
In recent years, deep learning models
have been widely applied to sentence-level RE and achieved remarkable success \cite{han2018hierarchical,yu2019beyond}.

Despite the great success of previous work, sentence-level RE suffers from a serious restriction in practice: a large amount of relational facts are expressed in multiple sentences.
Taking Figure 1 as an example, in order to identify the relational fact (\emph{Chris Carter}, \emph{creator}, \emph{Fox Mulder}), one should first identify the fact "Nisei" is an episode of the American science fiction television series from sentence 1, then identify the facts that \emph{Fox Mulder} is a character in "Nisei" and \emph{Chris Carter} is one of the writers of "Nisei" from sentence 8 and 3 respectively.
To extract these relational facts, it is necessary to infer and aggregate over multiple sentences.
Obviously, most traditional sentence-level RE models often fail to generalize extraction to this situation.
To move RE forward from sentence level to document level, many efforts have been made \cite{wang2019fine,Yao2019DocREDAL},
but most previous methods used only entity-level information and this is not adequate.
Thus, there are still some deep-seated problems unsolved in document-level RE.

To predict the relation between two entities, we argue that the document-level RE model requires taking advantage of multi-granularity inference information: entity level, sentence level and document level.
Let’s go back to the former example, 
entity-level inference information is derived from the semantic of all mentions of \emph{Chris Carter} and \emph{Fox Mulder} in the document, 
sentence-level inference information represents the information related to relational facts in each sentence,
document-level inference information aggregates all the necessary information in supporting sentences (sentence 1, 3 and 8) and discards information in noise sentences.
Technically, it is clear that document-level RE faces two main challenges: 
(1) How to obtain the inference information with different granularity; 
(2) How to aggregate these different granularity inference information and make the final prediction.

In this paper, we propose a new neural architecture, Hierarchical Inference Network (HIN), to tackle above challenges.
Specifically, inspired by translation constraint \cite{bordes2013translating}, which models a relational fact $r({e}_{h}, {e}_{t})$ with $e_{h} + r \approx e_{t}$,
we apply this translation constraint to target entity pair.
Besides, a bi-affine layer is also used to obtain bilinear representation for the target entity pair.
To jointly attend to information from different representation subspaces, we implement the above two transformations in multiple subspaces in parallel, and acquire entity-level inference information.
To obtain the sentence-level inference information, we first apply vanilla attention mechanism to calculate the vector representation for each sentence, which enables our model to pay more attention to the informative words.
Then we adopt the semantic matching method which is widely used in natural language inference (NLI) domain to compare the entity-level inference information with each sentence vector. 
Furthermore, in order to calculate the document-level inference information, we apply a hierarchical BiLSTM and again use attention mechanism to distinguish crucial sentence-level inference information for overall document-level inference representation.
Finally, we aggregate inference information of different granularity, the entity-level and document-level inference representations are combined into a fixed-length vector, 
which is further fed into a classification layer for prediction.

To summarize, we make the following contributions:

1. We propose a Hierarchical Inference Network (HIN) for document-level RE, which is capable of aggregating inference information from entity level to sentence level and then to document level.

2. We conduct thorough evaluation on DocRED dataset. Results show that our model achieves the state-of-the-art performance. We further demonstrate that using BERT representations further substantially boosts the performance.

3. We analyze the effectiveness of our model on different number of supporting sentences and experimental results show that our model performs much better than previous work when the number of supporting sentences is large.

\section{Task Description }
For document-level RE, the input is a document with annotated entities, as well as multiple occurrences of each entity, i.e., entity mentions,
the goal is to identify all the related entity pairs in the document.
Following \cite{Yao2019DocREDAL}, we transform RE into a classification problem.
We use upper case letters to represent entities $\small \left(E_{1}, \cdots, E_{m}\right)$ and lower case letters to represent mentions $\small \left(e_{1}, \cdots, e_{m}\right)$.
The RE model is given a relation candidate $\small \left(E_{a} , E_{b}, D\right)$ and expected to output the relations between $\small E_{a}$ and $\small E_{b}$, where $\small E_{a}$ and $\small E_{b}$ are entities in the document $\small D$.
\begin{figure*}[ht]
	\centering
	\includegraphics[scale=0.40]{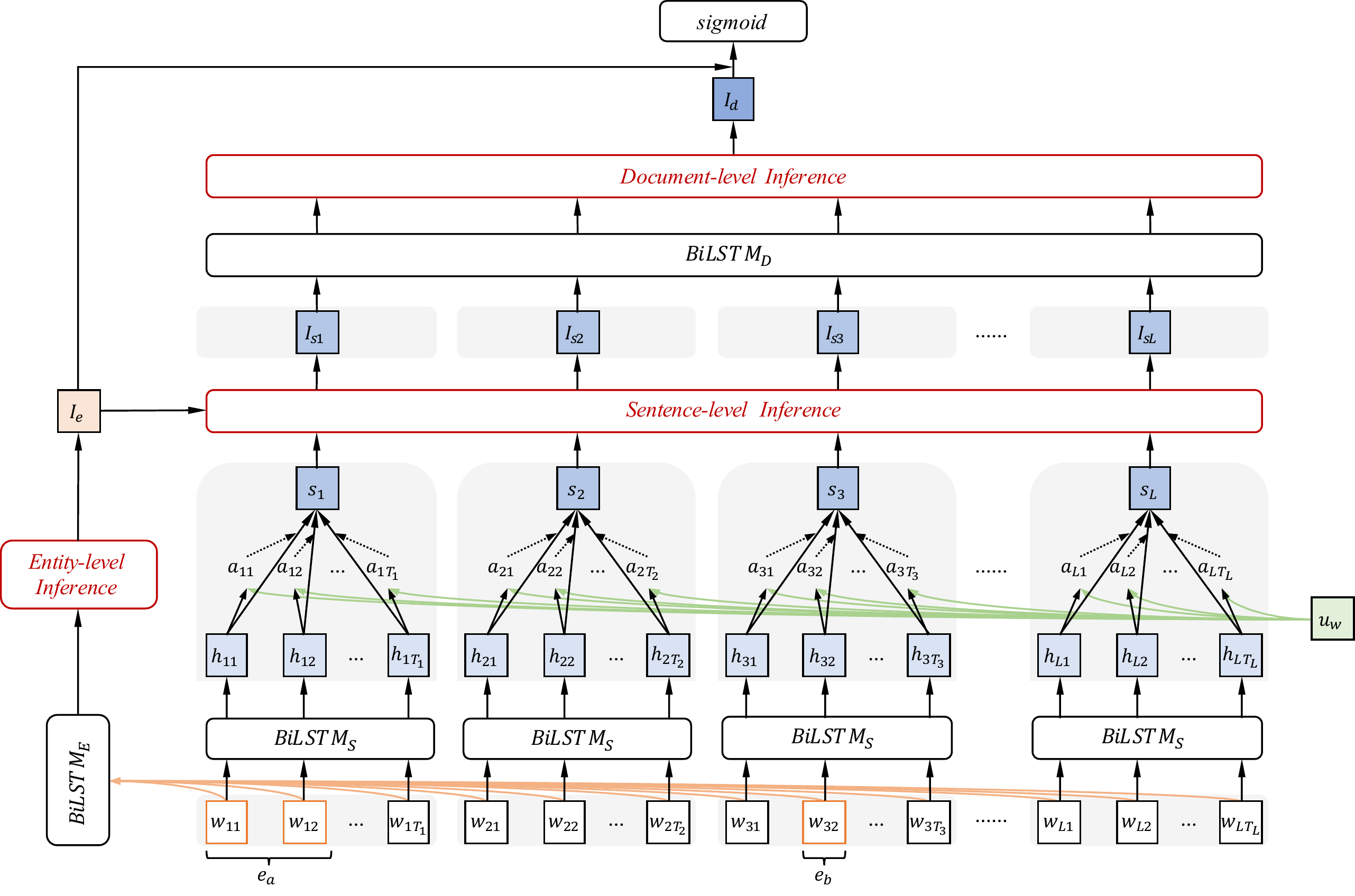} 
	\protect\caption{The overall architecture of the Hierarchical Inference Network (HIN)}
	\label{Fig. 2}
\end{figure*}
\section{Proposed Approach}
Figure 2 gives an illustration of our model.
We describe the details of different components in the following sections. 

\subsection{Input Layer}
\begin{itemize}
	\item 
	\textbf{Word Embeddings}
	In order to capture the meaningful semantic information of words, we map each word into a low-dimensional word embedding vector. 
	The dimension of word embeddings is $\small d_{w}$.
	
	\item 
	\textbf{Entity Type Embeddings}
	We utilize the entity type information to enrich the representation of the input.
	The entity type embedding is obtained by mapping the entity type (e.g., PER, LOC, ORG) into a vector. 
	The dimension of entity type embeddings is $\small d_{t}$.
	
	\item 
	\textbf{Coreference Embeddings}
	Usually each entity may be mentioned many times in a document.
	Following previous work, we assign entity mentions corresponding to the same entity with the same entity id, which is determined by the order in which entities appear in the document.
	Then entity ids are embedded into vectors. 
	The dimension of coreference embeddings is $\small d_{c}$.
\end{itemize}

We concatenate all three embeddings together for each word $\small {w}_{i}$, and a document is transformed into a matrix $\small \textbf{X}=\left[\textbf{w}_{1}, \textbf{w}_{2}, \ldots, \textbf{w}_{{n}}\right]$, 
where each word vector $\small \textbf{w}_{i} \in \mathbb{R}^{d_{w}+ d_{t}+d_{c}}$ and $n$ is the length of the document.

\subsection{Entity-Level Inference Module}
In this section, we compute the entity-level inference information for target entity pair.
To represent each word in its context, we encode the document $\small \textbf{X}=\left\{\textbf{w}_{i}\right\}_{i=1}^{n}$ 
into a hidden state vector sequence $\small \left\{\textbf{h}_{i}\right\}_{i=1}^{n}$ with bi-directional LSTM:
\begin{equation}
\small
\textbf{h}_{i}=\mathrm{BiLSTM}_{E}\left(\textbf{w}_{i}\right), i \in[1, n].
\end{equation}
where $\small \textbf{h}_{i} \in \mathbb{R}^{d}$ is a contextualized representation of $\small w_{i}$, summarizing the context information centered around $\small w_{i}$. 

Considering that an entity may be mentioned many times in a document and a mention may also contain more than one word, we represent each entity and mention with the average of the embeddings of different elements.
Correspondingly, the mention representation is formed as the average of the words that the mention contains, the entity representation is computed as the average of the mention representations associated with the entity:
\begin{equation}
\small
\textbf{e}_{l}=avg_{w_{i}\in e_{l}}(\textbf{h}_{i}), \quad 
\textbf{E}_{a}=avg_{e_{l}\in E_{a}}(\textbf{e}_{l})
\end{equation}
We claim that it is beneficial to allow the model to jointly attend to information from different representation subspaces, thus, we use different learnable projection matrices to project entities into  $K$ subspaces:
\begin{equation}
\small
\mathbf{E}_{a}^{k}= \mathbf{W}_{k}^{(1)}(ReLU(\mathbf{W}_{k}^{(0)} \, \mathbf{E}_{a}))
\end{equation}
where $\small \mathbf{E}_{a}^{k} \in \mathbb{R}^{k}$ corresponds to the representation of $\small E_{a}$ in the k-th latent space, 
$\small \mathbf{W}_{k}^{(0)} \in \mathbb{R}^{d\times d}$ and $\small \mathbf{W}_{k}^{(1)} \in \mathbb{R}^{d\times k}$ are the learnable projection matrices corresponding to the $\small k$-th subspace.
For each of these projected versions, we perform the entity-level inference in parallel.
These are concatenated and once again projected, resulting in the final entity-level inference information.

Inspired by TransE \cite{bordes2013translating} which modelled a triple $\small r({e}_{h}, {e}_{t})$ with $\small \textbf{e}_{h} + \textbf{r} \approx \textbf{e}_{t}$, we argue that $\small (\mathbf{E}_{b} - \mathbf{E}_{a})$ could represent the relation between $\small E_{a}$ and $\small E_{b}$ in the document to some extent.
In addition, a bilinear representation can be obtained by a bi-affine layer to enhance the expression ability of model.
We define the following formula as entity-level inference representation in the $k$-th latent space:
\begin{equation}
\small
\textbf{I}_{e}^{k}={Concat}\left(\textbf{E}_{a}^{k} \, \textbf{R}^{k} \, \textbf{E}_{b}^{k} ; \, \textbf{E}_{b}^{k} - \textbf{E}_{a}^{k} ; \, \textbf{E}_{a}^{k} ; \, \textbf{E}_{b}^{k}\right)
\end{equation}
where $\small \textbf{R}^{k}\in \mathbb{R}^{k \times k \times k}$ is a learned bi-affine tensor, $\small {Concat}$ denotes concatenation.

Moreover, we believe that the relative distances between two target entities can help us better judge the relations. 
Empirically, we use the relative distances between the first mentions of the two entities as the relative distances between two target entities. Finally, all entity-level inference representations in different latent space and the relative distance embeddings are fed into a feed-forward neural network (FFNN) to form the final entity-level inference information:
\begin{equation}
\small
\textbf{I}_{e}=G_{e}\left(\left[\textbf{I}_{e}^{1} ; \, ... ; \, \textbf{I}_{e}^{K} ; \, \textbf{M}\left(d_{b a}\right) - \textbf{M}\left(d_{a b}\right) \right]\right)
\end{equation}
here $\small G_{e}$ is a FFNN with ReLU activation function, 
$\small \textbf{M}$ is an embedding matrix, $\small d_{a b}$ and $\small d_{b a}$ are the relative distances between $\small E_{a}$ and $\small E_{b}$ in the document.
$\small \textbf{I}_{e} \in \mathbb{R}^{d}$ describes relation features between $\small E_{a}$ and $\small E_{b}$ at entity level. 

\subsection{Hierarchical Document-Level Inference Module}
In this section, we propose a hierarchical inference mechanism, inference information is aggregated from entity level to sentence level and then to document level.
In this way, our model can aggregate all useful information of the document.

\subsubsection{Sentence-Level Inference}
Assume that a document contains $\small L$ sentences, and
$\small w_{j t}$ represent the $\small t$-th word in the $\small j$-th sentence.
Given the $\small j$-th sentence $\small S_{j}$, to represent words in its context, the sentence is fed into a BiLSTM encoder:
\begin{equation}
\small
\textbf{h}_{j t}=\mathrm{BiLSTM}_{S}\left(\textbf{w}_{j t}\right), t \in[1, T_{j}].
\end{equation}
Since different words in a sentence are differentially informative, inspired by \cite{yang2016hierarchical}, we introduce the vanilla attention mechanism to enable our model to selectively assign higher weights for the informative words and lower weights for the other words.
Then we aggregate the representations of those informative words to form a sentence vector. Specifically,
\begin{align}
\small
\alpha_{j t}&=\textbf{u}_{w}^{\top} \tanh \left(\textbf{W}_{w} \textbf{h}_{j t}+\textbf{b}_{w}\right) \\
a_{j t}&=\frac{\exp \left(\alpha_{j t}\right)}{\sum_{t} \exp \left(\alpha_{j t}\right)} \\
\textbf{S}_{j}&=\sum_{t} a_{j t} \textbf{h}_{j t}
\end{align}
where $\small \textbf{u}_{w}, \textbf{b}_{w} \in \mathbb{R}^{d}$ and $\small \textbf{W}_{w} \in \mathbb{R}^{d \times d}$ are learnable parameters.
Word hidden state $\small \textbf{h}_{j t} \in \mathbb{R}^{d}$ is first fed through a one-layer MLP,
then we obtain weights of words by measuring ``which words are more related to the target entities".
Finally, we compute the sentence vector $\small \textbf{S}_{j}$ as a weighted sum of the word hidden states. 

For obtaining the sentence-level inference information, we adopt a semantic matching method which is used in previous NLI model \cite{chen2016enhanced}. Through comparing sentence vector $\small \textbf{S}_{j}$ with entity-level inference representation $\small \textbf{I}_{e}$, we can derive sentence-level inference representation $\small \textbf{I}_{sj}$ for the $\small j$-th sentence:
\begin{equation}
\small
\textbf{I}_{sj}=G_{s}\left(\left[\textbf{S}_{j} ; \, \textbf{I}_{e} ; \, \textbf{S}_{j}-\textbf{I}_{e} ; \, \textbf{S}_{j} \circ \textbf{I}_{e}\right]\right).
\end{equation}
where $\small G_{s}$ is FFNN with ReLU function, a matching trick with elementwise subtraction and multiplication is used for building better matching representations \cite{Mou2015NaturalLI}.
$\small \textbf{I}_{sj}$ represents the inference information derived from the $\small j$-th sentence.

\subsubsection{Document-Level Inference}
In order to distinguish crucial sentence-level inference information for overall document-level inference representation,
vanilla attention  mechanism is again used.
We build a BiLSTM followed by the attention network on top of the sentence-level inference vectors ($\small \textbf{I}_{s}$) to aggregate all essential evidence information scattered in different sentences:
\begin{align}
\small
\textbf{c}_{s j}&=\mathrm{BiLSTM}_{D}\left(\textbf{I}_{s j}\right), j \in[1, L] \\
\alpha_{j}&=\textbf{u}_{s}^{\top} \tanh \left(\textbf{W}_{s} \textbf{c}_{s j}+\textbf{b}_{s}\right) \\
a_{j}&=\frac{\exp \left(\alpha_{j}\right)}{\sum_{j} \exp \left(\alpha_{j}\right)} \\
\textbf{I}_{d}&=\sum_{t} a_{j} \textbf{c}_{s j}	
\end{align}
here $\small \textbf{u}_{s}, \textbf{b}_{s} \in \mathbb{R}^{d}$ and $\small \textbf{W}_{s} \in \mathbb{R}^{d\times d}$ are learnable parameters,
$\small \textbf{I}_{d} \in \mathbb{R}^{d}$ is the document-level inference representation which represents all the inference information that we can obtain from the document.

\subsection{Prediction Layer}
To better integrate inference information of different granularity, 
we concatenate entity-level inference representation $\small \textbf{I}_{e}$ and document-level inference representation $\small \textbf{I}_{d}$ together to form the final inference representation.
Since there are often multiple relations holding between an entity pair, we use a FFNN with the $\small sigmoid$ function to calculate the probability of each relation:
\begin{equation}
\small
P\left(r | E_{a}, E_{b}\right)=sigmoid \left(\textbf{W}_{r} \left[\begin{array}{l}{\textbf{I}_{e}} \\ {\textbf{I}_{d}}\end{array}\right]+\textbf{b}_{r}\right).
\end{equation}
where $\small \textbf{W}_{r}$, $\textbf{b}_{r}$ are the weight matrix and bias for the linear transformation.

A binary label vector $\small \textbf{y}$ is set to indicate the set of true relations holding between the entity pair, where 1 means an relation is in the set, and 0 otherwise. 
In our experiments, we use the binary cross entropy (BCE) as training loss:
\begin{equation}
\small
Loss = -\sum_{r=1}^{l} y_{r} \log \left(p_{r}\right)+\left(1-y_{r}\right) \log \left(1-p_{r}\right).
\end{equation}
where $\small y_{r} \in\{0,1\}$ is the true value on label $\small r$ and $\small l$ is the number of relations.

Given a document, we rank the predicted results by their confidence and traverse this list from top to bottom by F1 score on dev set, the probability value corresponding to the maximum F1 is picked as threshold $\small \delta$.
This threshold is used to control the number of extracted relational facts on test set.

\section{Experiments}
\subsection{Dataset}

To evaluate the effectiveness of our model, we use the DocRED dataset \cite{Yao2019DocREDAL}, which is the largest human-annotated document-level RE dataset constructed from Wikidata and Wikipedia.
DocRED contains over 5,053 documents, 40,276 sentences, 132,375 entities and 96 frequent relation types.
Entity types in DocRED are annotated.
It is also introduced by the author of DocRED that about 40.7\% of relational facts can only be extracted from multiple sentences and 61.1\% relational instances require a variety of reasoning.

\begin{table}[t]
	\small
	\caption{Performance of different models on DocRED (\%).}
	\centering
	\label{tab3}
	\setlength{\tabcolsep}{2mm}{
		\begin{tabular}{l|c c|c c}
			
			\hline
			\multirow{2}{*}{\textbf{Model}}&\multicolumn{2}{c|}{\textbf{Dev}}&\multicolumn{2}{c}{\textbf{Test}} \\
			~& \textbf{Ign F1} & \textbf{F1} & \textbf{Ign F1} & \textbf{F1}\\
			\hline
			
			CNN-RE \cite{Yao2019DocREDAL} & 41.58 & 43.45 & 40.33 & 42.26 \\
			LSTM-RE \cite{Yao2019DocREDAL} & 48.44 & 50.68 & 47.71 & 50.07 \\
			BiLSTM-RE \cite{Yao2019DocREDAL} & 48.87 & 50.94 & 48.78 & 51.06 \\
			Context-Aware \cite{sorokin2017context} & 48.94 & 51.09 & 48.40 & 50.70 \\
			\hline
			\textbf{HIN-GloVe} & \textbf{51.06} & \textbf{52.95} & \textbf{51.15} & \textbf{53.30} \\
			\hline
			\hline
			BERT-RE \cite{wang2019fine} & - & 54.16 & - & 53.20 \\
			BERT-Two-Step \cite{wang2019fine} & - & 54.42 & - & 53.92 \\
			\hline
			\textbf{HIN-BERT} & \textbf{54.29} & \textbf{56.31} & \textbf{53.70} & \textbf{55.60} \\
			\hline
			\hline
			
	\end{tabular}}
\end{table}

\subsection{Comparison Models \& Evaluation Metrics}
We compare our model against the following  document-level RE baselines:

\textbf{CNN/LSTM/BiLSTM-RE:}  
They first encode a document into a hidden state vector sequence with CNN/LSTM/BiLSTM as encoder, and then predict relations for each entity pair by feeding them into a bilinear function \cite{Yao2019DocREDAL}.

\textbf{Context-Aware:}
It uses an LSTM-based encoder to jointly learn representations for all relations in the context, and then combines other context relations with target relation to make the final prediction \cite{sorokin2017context}.

\textbf{BERT-RE:}
It uses BERT to encode the document, entities are represented by their average word embedding.
A BiLinear layer is applied to predict the relation between entity pairs \cite{wang2019fine}.

\textbf{BERT-Two-Step:}
Based on BERT-RE, it models the document-level RE through a two-step process. The first step is to predict whether or not two entities have a relation, the second step is to predict the specific relation \cite{wang2019fine}.

\textbf{HIN:}
This is the main model of this paper. Multi-granularity inference information is used to better model complex interactions between entities.

The widely used metric F1 is used in our experiments. Moreover, since some relational facts present in both training and dev/test sets, we also report the F1 excluding those relational facts and denote it as Ign F1.

\subsection{Implementation Details}
We try two embedding methods in our experiments: 100-dimensional GloVe \cite{pennington2014glove} embeddings and BERT representations \cite{devlin2018bert}.
For the BERT representations, the base uncased English model with dimension 768 is used, we map word representations into 100 dimensional vectors by a linear projection layer.
Once the word representations are initialized, they are fixed during training.
The embedding dimensions of coreference, distance and entity type are all set to be 20.
For LSTM encoder, the dimension of the hidden units is 128.
The number of latent space is 2.
Furthermore, we regularize our network using dropout and the dropout ratio is 0.2.
We optimized our model using Adam \cite{kingma2014adam}, with learning rate of $10^{-4}$, $\beta_{1}=0.9$, $\beta_{2}=0.999$.
The batch size is set to be 12 and the value of threshold $\delta$ is determined by the performance on the dev set.

\subsection{Experimental Results and Analyses}
\subsubsection{Overall Performance}
Experimental results are shown in Table 1. From the results, we can observe that:
(1) Compared with BiLSTM-RE, the state-of-the-art model without BERT, our HIN-GloVe achieves significant improvements of 2.24\% in F1, we claim that it is mainly due to the reasoning mechanism and hierarchical aggregation structure in HIN, which will be further discussed in ablation study.
(2) Even though BERT based models provides strong prediction power, HIN-BERT consistently improves over them,
which further proves the effectiveness of our hierarchical inference network.
(3) Although Context-Aware model combines context relations with the target relation, it can't use the evidence information in document as effectively as HIN. Hence our model also outperforms it by 2.60\% in F1.
(4) BERT representations further boost the performance of our model, the HIN-BERT approach outperforms all these previous methods, which indicates the importance of prior knowledge.

\begin{table}[t]
	\small
	\caption{Results of ablation study (\%).}
	\centering
	\label{tab4}
	\setlength{\tabcolsep}{5.2mm}{
		\begin{tabular}{l|c c}
			\hline
			\multirow{2}{*}{\textbf{Setting}}&\multicolumn{2}{c}{\textbf{Dev}} \\
			~& \textbf{Ign F1} & \textbf{F1} \\
			\hline
			\textbf{HIN-BERT} & \textbf{54.29} & \textbf{56.31} \\
			\hline
			- Translation mechanism & 53.09 & 55.10 \\
			
			- Bilinear transformation & 52.15 & 54.29 \\
			
			- Multispace & 52.44 & 54.59 \\
			
			- Sentence inference & 52.82 & 55.06 \\
			
			- Hierarchical aggregation & 51.36 & 53.50 \\
			
			- Above all & 49.95 & 52.10 \\
			\hline
	\end{tabular}}
\end{table}

\subsubsection{Ablation Study}
To study the contribution of each component in HIN-BERT, we run an ablation study on DocRED dev set (see Table 2). From these ablations, we find that:
(1) When we remove the translation mechanism and bilinear transformation, 
F1 score drops by  1.21\% and 2.02\% respectively,
which indicates that these two transformations can enhance the expression ability of HIN at the entity level.
(2) Removing the multi-space projection hurts the result by 1.72\%, which proves that it is beneficial to allow the model to jointly attend to information from different representation subspaces.
(3) F1 drops by 1.25\% when we remove the sentence-level inference mechanism, i.e., replacing the sentence-level inference vector with sentence vector. 
(4) F1 drops by 2.81\% when we discard the hierarchical aggregation approach. Instead, we run BiLSTM followed by mean-pooling layer over the whole document to get the document vector. 
(5) We also observe that F1 drops by 4.21\% when we discard the above all factors together.
In summary, all components play an important role in our model.
\begin{figure}[t]
	\centering
	\includegraphics[scale=0.50]{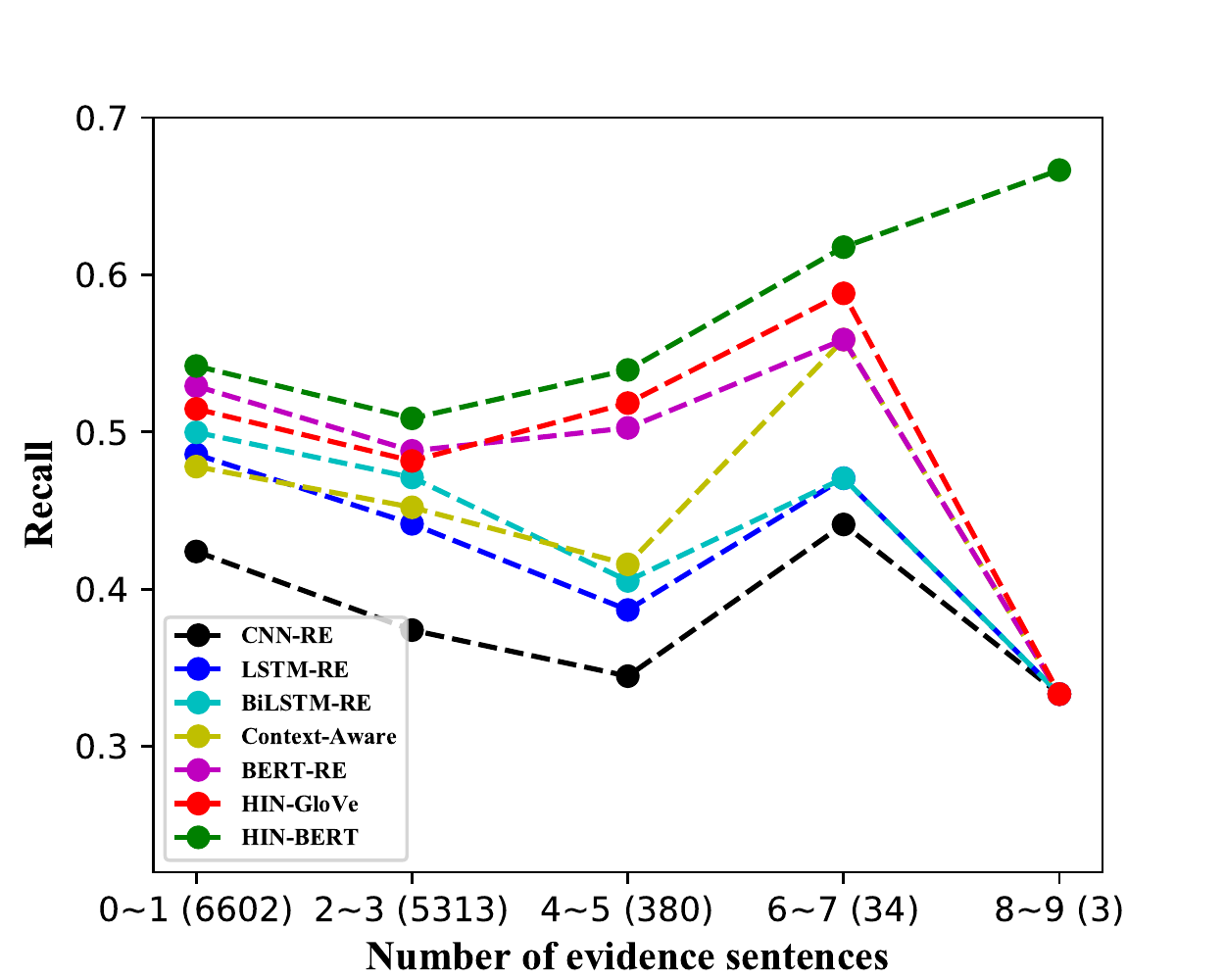}
	\protect\caption{Recall of models on relational facts with different number of supporting sentences. Numbers in parentheses represent the number of relational facts with different number of supporting sentences in dev set.}
	\label{Fig. 3}
\end{figure}
\subsubsection{Analysis by the number of supporting sentences}
As we discussed before, it is challenging for document-level RE to reason from multiple sentences.
To further prove the effectiveness of HIN, we analyze the recall on relational facts with different number of supporting sentences here.\footnote{Since there is no official code for BERT-Two-Step, its results are not counted.}
As shown in Figure 3, we find that our model always performs better than other baselines, especially when the number of supporting sentences increases gradually.
More specifically, HIN-GloVe even outperforms BERT-RE when the number of supporting sentences exceeds 4, 
which fully proves the superiority of HIN.
Note that when the number of supporting sentences exceeds 7, HIN-GloVe and other baselines behave the same.
We think this is because there are very few samples with more than 7 supporting sentences in dev set.
We believe when the number of relational facts with more supporting sentences increase our model will achieve better results.
\begin{table*}[t]
	\small
	\caption{The results predicted by BERT-RE and HIN-BERT.
		The reasoning type of each example is different and the first row for each example is the input document. The {\color{blue}{\textbf{\emph{head}}}}, {\color{cyan}{\textbf{\emph{tail}}}}, {\color{magenta}{\textbf{\emph{relation}}}} and \textcolor[rgb]{1.0, 0.44, 0.37}{\textbf{supporting sentences}} are colored accordingly.}
	\begin{center}
		
		\begin{tabular}{p{0.13\columnwidth} p{0.87\columnwidth}}
			
			\toprule
			\hline
			
			Logical \qquad  reasoning & 
			\textcolor[rgb]{1.0, 0.44, 0.37}{\textbf{[0]}} The Galaxy S series is a line of Samsung Electronics, a division of {\color{cyan}{\textbf{\emph{Samsung}}}}
			\textcolor[rgb]{1.0, 0.44, 0.37}{\textbf{[2]}} Galaxy S line has ... being {\color{cyan}{\textbf{\emph{Samsung}}}} 
			's flagship smartphones. 
			\textcolor[rgb]{1.0, 0.44, 0.37}{\textbf{[4]}} the latest smartphones in  Galaxy S series are the {\color{blue}{\textbf{\emph{Samsung Galaxy S9}}}} ... \\
			
			\hline
			\textbf{Relation} & \tabincell{l}{
				\textbf{Lable}: {\color{magenta}{\textbf{\emph{manufacturer}}}} \,
				\textbf{BERT-RE}: {\color{magenta}{\textbf{\emph{None}}}} \,
				\textbf{HIN-BERT}: {\color{magenta}{\textbf{\emph{manufacturer}}}}  	
			}\\
			
			\hline
			\hline
			
			Coreference reasoning & 
			\textcolor[rgb]{1.0, 0.44, 0.37}{\textbf{[0]}} {\color{blue}{\textbf{\emph{Robert Kingsbury Huntington}}}}, was a naval aircrewman and member of Torpedo Squadron 8.
			[2] ... {\color{blue}{\textbf{\emph{Huntington}}}} 
			was shot down during the Battle of Midway ...
			\textcolor[rgb]{1.0, 0.44, 0.37}{\textbf{[3]}} \underline{He} was born in {\color{cyan}{\textbf{\emph{Los Angeles}}}} , California ...\\
			
			\hline
			\textbf{Relation} & \tabincell{l}{
				\textbf{Lable}: {\color{magenta}{\textbf{\emph{birth place}}}} \,
				\textbf{BERT-RE}: {\color{magenta}{\textbf{\emph{death place}}}} \,
				\textbf{HIN-BERT}: {\color{magenta}{\textbf{\emph{birth place}}}}	
			}\\
			
			\hline
			\hline
			
			Common-sense reasoning &
			\textcolor[rgb]{1.0, 0.44, 0.37}{\textbf{[0]}} IBM Research – Brazil is one of twelve research laboratories comprising IBM Research , its first in {\color{cyan}{\textbf{\emph{South America}}}}.
			\textcolor[rgb]{1.0, 0.44, 0.37}{\textbf{[1]}} It was established in June 2010 , with locations in {\color{blue}{\textbf{\emph{São Paulo}}}} and Rio de Janeiro
			...\\
			
			\hline
			
			\textbf{Relation} & \tabincell{l}{
				\textbf{Lable}: {\color{magenta}{\textbf{\emph{continent}}}} \, \,
				\textbf{BERT-RE}: {\color{magenta}{\textbf{\emph{country}}}} \, \, \, \,
				\textbf{HIN-BERT}: {\color{magenta}{\textbf{\emph{country}}}}
			}\\
			
			\hline
			
			\bottomrule
			
		\end{tabular}
		\label{tab5}
	\end{center}
\end{table*}
\subsubsection{Case Study}
We compare our model with BERT-RE on some cases from dev set, as shown in Table 3.
(1) Example 1 represents the situation that logical reasoning is required.
Specifically, in order to identify the relational fact,
we have to first identify the fact that \emph{Galaxy S} series is a line of \emph{Samsung} from sentence 0 and 2,
then identify the fact \emph{Samsung Galaxy S9} is the latest smartphones in the Galaxy S series from sentence 4.
We explain that our model uses a hierarchical aggregation approach to collect inference information from multiple sentences, so that it can better deal with this complex inter-sentence relationship.
(2) Example 2 represents the case of coreference reasoning.
In this situation, we claim that the attention and reasoning mechanisms in sentence-level inference module can help us to identify that "He" refers to \emph{Robert Kingsbury Huntington} in sentence 3.
In the end, our model can identify the right relation while BERT-RE mistakenly assumes that \emph{Los Angeles} is the place where \emph{Robert Kingsbury Huntington} died.
(3) Example 3 is a case that needs to combine context information with common-sense knowledge.
Through some external common-sense knowledge, we might know that \emph{South America} is a continent and \emph{São Paulo} is a city, which is the useful information to help judge their relation.
We think the problem can be solved by adding some external knowledge and we leave it as our future work.

\section{Related Work}
In recent years, more and more neural models have been applied to RE.
Zeng el al. \cite{Zeng2014RelationCV} employed a one-dimensional CNN with additional lexical features to encode relations.
Miwa et al. \cite{miwa2016end} used LSTM with tree structures for RE.
Zhou el al. \cite{zhou2016attention} 
showed that combining CNN/RNN with attention mechanism can further improve performance.
And the emergence of various optimization algorithms \cite{li2015lingo,li2017riemannian,li2017learning} makes these neural models more effective.
Most existing RE work focuses on modeling within a single sentence.
However, usually documents provide more information than sentences.
Moving research from sentence level to document level is necessary.
Recently, there has been increasing interest in document-level RE.
Yao et al. \cite{Yao2019DocREDAL} 
proposed a large-scale human-annotated document-level RE dataset, DocRED, 
and first compute the representations for all entities then predict relations for each entity pair by feeding them into a bilinear function.
Wang et al. \cite{wang2019fine} used BERT to encode the document, it also used bilinear layer to predict the relation between entity pairs,
but it modelled the document-level RE through a two-step process. 
Most previous methods used only entity-level information and this is not adequate. In this paper, we propose to effectively aggregate the inference information of different granularity. 

\section{Conclusion}
In this paper, we proposed a Hierarchical Inference Network (HIN) for document-level RE.
It uses a hierarchical inference method to aggregate the inference information of different granularity: entity level, sentence level and document level.
We show that our method achieves state-of-the-art performance on the largest human-annotated DocRED dataset.
Experimental analysis shows that both the inference mechanism and hierarchical aggregation approach in our model play an important role.
In the future, we plan to incorporate external knowledge to further improve the proposed model.

\section{Acknowledgements}

This research is supported by the National Key Research and Development Program of China (No.2018YFB1004703).

%
%
%
\bibliographystyle{splncs04}
%
\bibliography{pakdd20}
\end{document}